\documentclass[twocolumn,superscriptaddress]{revtex4}

\usepackage{graphicx,epsfig}
\usepackage{amsmath}
\usepackage{amsfonts}

\usepackage{fancyhdr}
\usepackage[T1]{fontenc}
\usepackage[utf8]{inputenc}
\usepackage{textcomp}

\usepackage{siunitx}
\usepackage{eurosym}
\DeclareSIUnit{\EUR}{\text{\euro}}

\usepackage[english]{babel}

\usepackage{natbib}

\begin{document}



\pagestyle{fancy}
\lhead{\bf PSD2 Explainable AI Model for Credit Scoring}
\rhead{Neus Llop Torrent et al.}
\rhead{}
\rfoot{Bologna, September 2020}

\title{PSD2 AI Explainable Model for Credit Scoring}
\author{Neus Llop Torrent}
\affiliation{Politecnico di Milano, Graduate School of Business, Milan, Italy }
\affiliation{CRIF S.p.A, via Mario Fantin 1-3 Bologna, Italy }

\author{Giorgio Visani}
\affiliation{University of Bologna, School of Informatics \& Engineering}
\affiliation{CRIF S.p.A, via Mario Fantin 1-3 Bologna, Italy }

\author{Enrico Bagli}
\affiliation{CRIF S.p.A, via Mario Fantin 1-3 Bologna, Italy }


\begin{abstract}
{\bf Abstract:} The aim of this project is to develop and test advanced analytical methods to improve the prediction accuracy of Credit Risk Models, preserving at the same time the model interpretability. In particular, the project focuses on applying an explainable machine learning model to bank-related databases. The input data were obtained from open data. Over the total proven models, CatBoost has shown the highest performance. The algorithm implementation produces a GINI of 0.68 after tuning the hyper-parameters. SHAP package is used to provide a global and local interpretation of the model predictions to formulate a human-comprehensive approach to understanding the decision-maker algorithm. The 20 most important features are selected using the Shapley values to present a full human-understandable model that reveals how the attributes of an individual are related to its model prediction.

\end{abstract}

\maketitle


\section{Introduction}
Credit Scoring is defined as the set of decision models and their underlying techniques that support lenders evaluate consumer credit \cite{thomas2017credit}. Hence,  is the use of statistical methods normally adopted by banks and financial entities to estimate the likelihood that a loan applicant will or will not default \cite{gup2005commercial}. 
Credit score is usually developed starting from factors such as the payment history, type and time of the credit application, outstanding debt and length of credit \cite{barron2003value}.

New online ordering facilities, innovative payment mechanisms, friendlier online platforms and the improvement of user experience on e-commerce, triggered the expansion of credit card usage and accounts creation. Based on the Global Payments Cards Data and Forecasts to 2024 made by  Retail Banking Research(RBR), in 2018 cashless payments increased by an 18\% being payments cards the biggest contributors, accounting for the 57\%. Besides, while debit card payments grow 6\%, check declined 7\%. 
The raise on cashless methods are resulting in a transformation of the facilities and conditions to access credit. One interesting case is Russia, where the Faster Payments System introduced in 2019 allows instant fund transfers via mobile phones number and QR codes. The results of these regulatory changes combined with new payment technologies like contactless (which increased a 25\% in 2018) were reported by RBR in the  Payment Cards Issuing and Acquiring Europe 2020 to be the key causes for the 9\% increment of card acceptance on 2018.  
As a consequence, the demand for automatized plans has drawn the attention of commercial banks, willing to find more accurate and fast techniques to track client's loan eligibility not to fall behind in the technological revolution of online payments after the PSD2 open banking revolution \cite{gomber2018fintech}. The failure of the implementation of new techniques could have damaging consequences for this sector if it is not able to modernize and implement the technology demanded by the customers.

On the other hand, transactions' accounts give an idea of the spending habits of the citizens, a highly relevant macroeconomic factor on systematic risk \cite{KHANDANI}.  The huge amount of data also exemplifies the wide window of decisions to which the customers are exposed. Credit card practices can reveal not only current life-styles but also expectations or preferences for their future way of life. Additionally, apart from a reflector of the reality of a customer group, which can be taken from the credit bureau data, transaction data can be displayed as a gate to identify inclinations that explain their desired life. The consumption behaviour and life-practices can be designated as an explanatory frame for client ambitions, threats and preferences which is a more genuine representation that their current state \cite{cclifestyle}.
Therefore, considering clients' irrationality, the study goes beyond customer's classification by trying to explain, to which extend, some variables can show a tendency of subjects not self-identifying to their social class by not behave accordingly to their predictable behaviour.

Most common scorecard methods implement the well-known Logistic Regression, which notably reduced the time of assessment of applicants. While logistic regression can identify the reasons behind the model choice, its major drawback is the incapability to capture the non-linearity correlation among features \cite{wang2015large}.
On the other hand, Machine Learning models have lately shown an increase in the prediction power for Credit Risk Modelling, although they do not provide reliable explanations for the scores they come up. That is a particularly delicate issue in CRM since it is a highly regulated field: the General Data Protection Regulation (GDPR), as well as the "Ethical Guidelines for trustworthy, AI" \cite{Ethical} and the Report from the "European Banking Authority" testify the care dedicated to such topics by the European Community \cite{AIandLaw}.

The Credit Scoring ecosystem made a 180 degrees shift when the Payment Services Directive 2 (PSD2) introduced a new regulatory framework for the Single Euro Payments Area. The main objectives of PSD2 are improving customer's protection and security while encouraging innovation and competition among the players on the payments industry by ensuring that all have the same accessibility to data. To do so, PSD-2 gives free access through bank's APIs to third parties,  providing them access to their client's accounts data. However, due to the unprecedented changes in legislation this directive accounted for, there are still no clear policies detailing the technologies to use, and the type of data banks are obligated to share. 
Serving its purpose of promoting entrepreneurship, PSD2 successfully accelerated the creation of Paytechs start-ups in Europe from 2018 to 2019, being the customer's cashless paying habits one of the drivers of its success \cite{polasik2020impact}. 

The new open banking era has also awakened the interest of Big Techs,  remaining in the spotlight for the moment to enter the financial services industry. Whereas there is no clue on when and how this may occur, Apple already launched its credit card in 2019 while Google may begin opening consumer bank accounts in 2021 \cite {10.1093/epolic/eiaa003}.

Currently, alternative credit scoring systems are treated as protected trade secrets, raising concerns about privacy and emphasising the lack of transparency in how data is being collected and used \cite{CreditReporting}. For processing these volumes of data in a reasonable amount of time, advanced AI and Big Data techniques are required. 
After the publication in April 2019 of the  {\it Ethical Guidelines for Trustworthy AI} by the European Commission, not being able to explain to a customer why he/she is not suitable to receive a loan has not been considered a valid solution anymore. Hence, to ensure AI transparency,  explainability has to be secured.
AI models explainability had drawn all the attention assisting the trustworthiness criteria. Therefore, before exploiting the Machine Learning potential in the Credit Scoring field, it is mandatory to address the interpretability issue. 
How reasonable is to make the machine learning 'black box' models responsible for its judgments when there is not a clear comprehension of the practice developed to reach their decisions? For instance, segregating by gender, location, origin, race may be inevitable without a clear understanding of the process followed by the algorithm \cite{CreditReporting}.

To manage this trade-off, our proposal concerns applying state of the art tools on top of a well-performing black-box algorithm. By doing so, we wish to retain the increased predictive power of Machine Learning, while providing meaningful explanations to the applicants involved, as well as to the regulator. The final goal of the work is to identify defaulted customers within their first 12-month relationship with the bank.

This problem was addressed by Jing Zhou from the Renmin University of China which proposed a method for creating  features for credit scoring focusing on the frequency, recentness and monetary value of the account information, using ML models \cite{huang2018rfms}. For historical transaction data, new time-series approaches based on Recurrent Neural Networks and LSTM are showing outstanding results, even they are still in a research phase \cite{LSTMarticle}. On the other hand, more classical approaches such as Logistic Regression, Decision Trees and GradientBoosting Classifiers are also implemented for more complex data sets \cite{yang2018comparison}.

In this paper, we will show how to create an explainable credit risk model based solely on open-data account transactions. We will discuss how we created the feature vectors from the raw data,  how the model operates and the reasoning behind the forecasted decisions. Furthermore, we will show how it is possible to interpret the model decisions, taking advantage of state-of-the-art explainability libraries, and we will discuss the model performances and behaviour. 

\section{Dataset Description}
The data used comes from an open-data database publicly available \cite{financialdatawebaddress}
Data contains account, transaction and client information.

\subsection{Data sources}

The four primary datasets collected four tables with differing information. The analysis will begin with a description of each table, followed by an explanation of how the distinct tables are combined to establish the final dataset. 

The credit history of the clients has been used to assess the creditworthiness of the clients. When a client has been insolvent it is considered a bad payer and the Performance variable is set to 1. In the opposite case, the Performance will be 0 meaning it is a good customer.

From  the accounts information, the account ID, the account balance, the date when the account balance was checked were used. The currency and account creation day were removed to be considerate irrelevant for the analysis. The account balance date will be used to back-propagate the account balance amount for all days from this date until the beginning of the data collection. 
The 4 variables in the transactions dataset are the transactions ID, the account ID from which had been performed the transaction and the date and amount of the transaction. 

\begin{figure}[h]
    \centering
    \includegraphics[width= \columnwidth ]{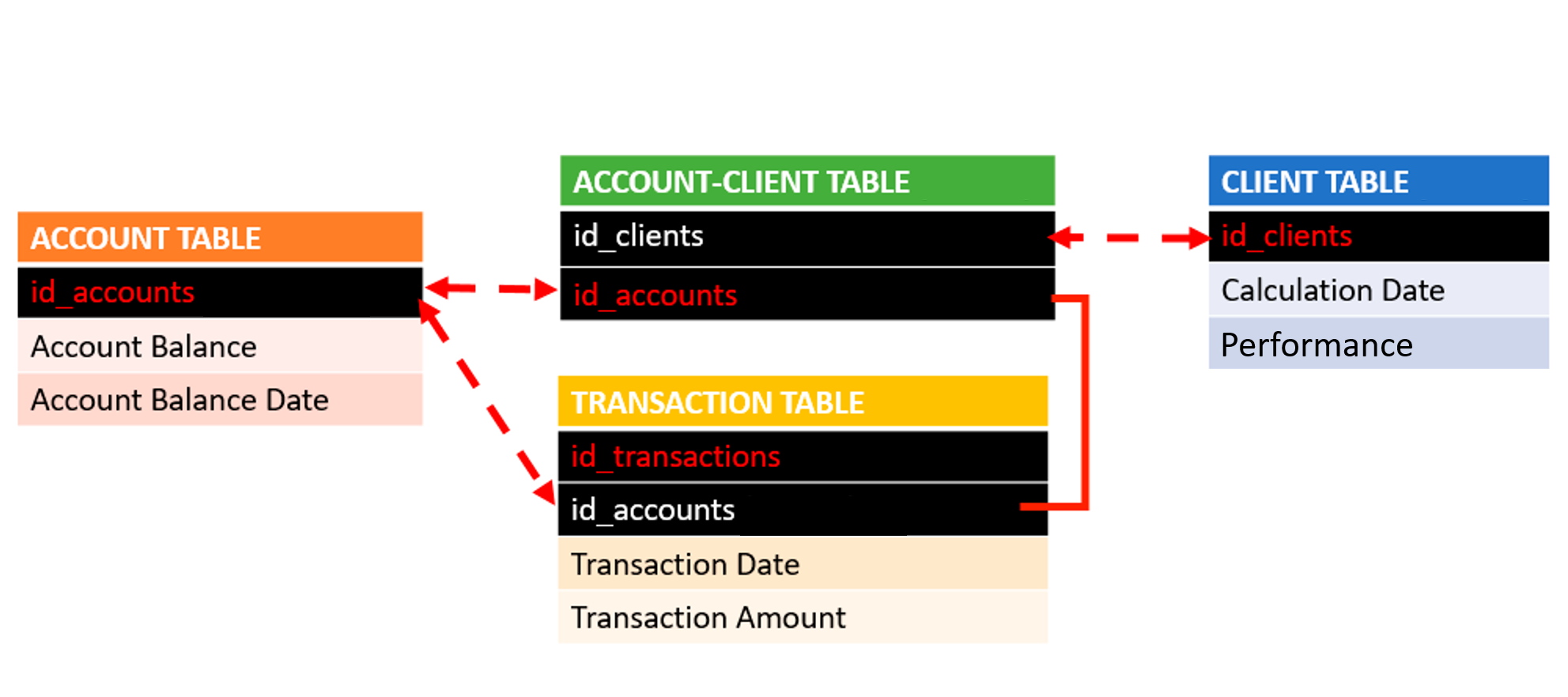}
    \caption{Scheme describing the main characteristics of the different tables and how they are related. Words coloured in red define the key of the table.Boxes in black identify the features used to join the tables.}
    \label{fig:table scheme}
\end{figure}

The aggregated dataset on { \bf Figure \ref{fig:table scheme}} contains the selected features from every table merged using the transaction id as the identification key. To relate the different tables a unique identifier is selected for each table. The account and client tables act as a link to merge the account table and the client table using the accounts and clients IDs. Ultimately, the transaction table is merged using the accounts ids.

\subsection{Data analysis}

\begin{figure}[h]
    \centering
    \includegraphics[ width=60mm ]{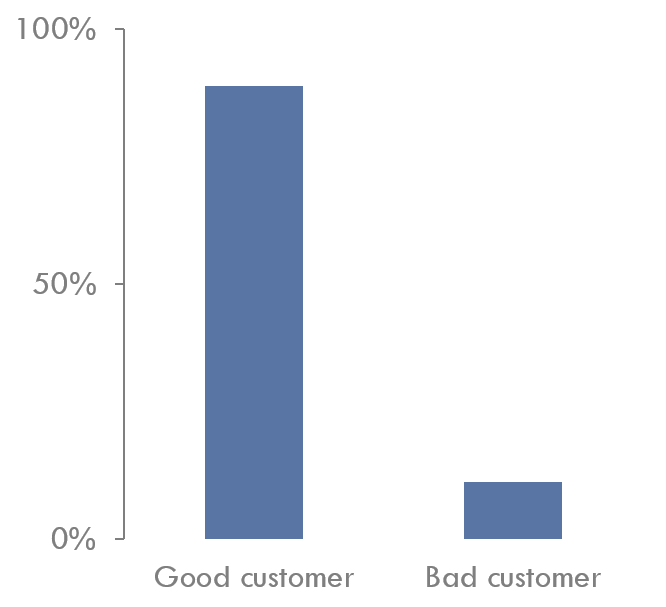}
    \caption{Histogram representing the distribution of bad and good customers along the dataset.}
    \label{fig:performance norm}
\end{figure}

From the total number of clients a 88.9 \% were rated Good clients while only a 11.1\% were bad as shown in {\bf Figure \ref{fig:performance norm}}. Consequently, as the target variable has more observations in one particular class than the other, it can be considered an unbalanced dataset which will influence the algorithm behaviour.
The presence of an unbalanced dataset may result in the over-fitting of the majority class as the model will tend to favour the majority class regardless of the input variables (This problem will be addressed in Section $III.C.2$). 

\begin{figure}[h]
    \centering
    \includegraphics[width= \columnwidth ]{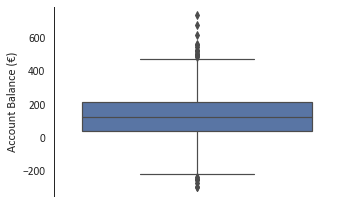}
    \caption{Box plot representing the distribution of customers' account balance.}
    \label{fig:account box plot}
\end{figure}

The account balance distribution is slightly shifted toward the positive balance as shown in {\bf Figure \ref{fig:account box plot}}. The maximum is found is $731$ \EUR  while the minimum account balance is $-300$ \EUR. 

The account-client relation is not bidirectional and unique. In fact, an important amount of accounts have more than a single client as it is also common to find clients with more than one account. 
 
\begin{figure}[h]
    \centering
    \includegraphics[width= \columnwidth ]{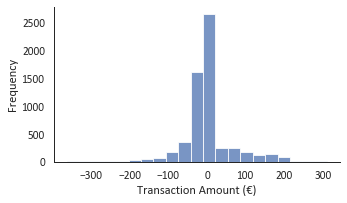}
    \caption{Histogram of the distribution of the transactions amount.}
    \label{fig:transaction histogram}
\end{figure}

The distribution of the transactions is shifted toward the negative side as shown in {\bf Figure \ref{fig:transaction histogram}}. The maximum is $+364$\EUR while the minimum transaction is $-364$\EUR. 

\section{Data Manipulation}

\subsection{Feature vectors construction}

We propose a learning algorithm capable of estimating the account's probability to default. Practically, this means it does not predict a good or bad customer but instead reports a likelihood. When the probability surpasses a proposed threshold set to $1/2$ by default, the account will be considered as bad. Interestingly, this popular cutoff is not backed by any theoretical justification as denoted by Eric Rosenberg \cite{rosenberg1994quantitative}. Further complicating the issue, these predictions are reported as point estimates (with its model's implied error).

First, transactions are used to get the account balance for each day. Second, the different KPIs will be created for a window of three months, considering the day the client asked for the credit as the end.

\subsection{Pre-processing and Feature Engineering}

This part addresses the processing stage necessary to handle the irregular temporal distribution of transactions for each account. In particular, the number of events varies for each customer and is not constant along the time. To control this, we built 3 months windows to establish a constant period. This is followed by a creation of variables through alternating the length and time location of the periods and computing aggregation operations using the transactions and account balance. As a result, 106 variables are obtained.  
After this process, all the accounts with the new shortened periods empty, will be removed from the dataset. 

\subsection{Feature Selection and Dimensionality Reduction}
Specific features may not be useful for modelling. The criteria we adopted to find those features were variables with: single values, a high percentage of missing values and high correlation. 
Zero features had one single value.
For all variables, a percentage of less than a 0.7\% of missing values was measured. Hence, no variables were removed based on this criteria. Both \emph{NaNs} and \emph{zeros} were also considerate missing values.
27 features were removed due to a correlation higher than 0.95\%. The collinearity between pairs of features was calculated using the Pearson correlation coefficient. For each pair, when the correlation was higher than the specified threshold, the second variable by order of appearance was removed. From the 106 variables, 79 were kept. 

After this first feature reduction, different models were run to obtain a benchmark. 

However, to have an interpretable model, a smaller set of features is needed. Hence, to proceed with the dimensionality selection, the SHAP values will be used to select the most important features \cite{lundberg2017unified}.
This condition cannot be reached until a model has been trained. To be consistent with our model selection we choose CatBoost as the benchmark algorithm. In the explainability section of the paper, we will cover how the variables are selected using SHAP.
It is just worth mentioning that the feature importance is a measure of the average impact the variable has on the model performance. Hence, by taking the mean of the absolute value of the SHAP values of the variable we can rank the variables by the influence they had on the prediction.

Based on the SHAP variable importance, we selected the 20 most important features for the model.
\subsubsection{Scaling}
Deep learning algorithms demand standard normal distributed data, which implies a \emph{mean} centred at zero $\mu = {\frac{1}{N} } {\sum_{i=1} ^N {(x_i)}}=0$ and a \emph{standard deviation} 
equal to one $ \sigma =  \sqrt{{\frac{1}{N} } \sum_{i=1} ^N (x_i - \mu)^2}=1$. The unit variance will restrict the model from having a preference for the features with a variance of higher orders of magnitude. Thus, we introduce a scaling {\bf \eqref{eq:1}} for the sample values of the neural network. For the remaining models, no normalization or standardization will be performed as these methods alter the relative distance between the feature's vectors and may impact on the forecast. 
\begin{equation}
z = {\frac{x-\mu}{\sigma}}
\label{eq:1}
\end{equation}

\subsubsection{Unbalanced Dataset}

As discussed in section two, there is an asymmetrical number of good /bad customers as the formers are largely more popular. Therefore, it can be studied as an unbalanced dataset problem. As the amount of bad customers accounts for the $11.1\%$ of the data, it may be reasonable to interpret it as an imbalanced dataset problem seeing the bad customers as a rare event. However, the nature of the problem makes it intuitively reasonable to consider it as an unbalanced problem, because, although it is less frequent the default of an account, it is still a widely spread event. Before considering any resampling method the data has to be split between the training and testing set as the method will be only applied for the training set. 

Hence, we proposed and tested a collection of 4 different resampling methods: undersampling the majority class, oversampling the minority class, the Synthetic Minority Oversampling Technique (SMOTE) and proportionally class weighting.

Randomly undersampling the majority class method consists of taking all the samples from the minority class and take from the majority class the same amount of samples. This method has the disadvantage that there is a significant quantity of data that will never be used to train the model. As a consequence, the model could underfit since important information from the majority class may have been lost \cite{more2016survey}. 
The opposite approach will be randomly oversampling the minority class. In this method instead of throwing part of the samples, the data from the minority class is replicated to reach the same size as the dominant class. However, as the samples of the non-dominating class are just a copy, they may be useless to gain knowledge from the data. 
Unlike oversampling, to solve the potential overfitting, SMOTE differentiates by creating new synthetically samples for the minority class using the nearest neighbours algorithm \cite{chawla2002smote}. Two improved variations of vanilla SMOTE will be explored. Baseline-SMOTE differentiates by focusing on the point that fall in the borderline between the classes. Then, it focuses on mainly taking the point of the minority class that was close to the border incorrectly classified and uses them for the synthetic duplication. That results in a better understanding of the region of conflict, bringing more knowledge where the algorithm finds difficulty in differentiating the classes \cite{han2005borderline}. A similar alternative is SVM-SMOTE, which follows the same idea but using a Super Vector Machine instead of a KNN to identify the points that fall in the frontier \cite{nguyen2011borderline}.

Finally, class-weights changes the weight of each class in the objective function. We will use \emph{class weight} by assigning a weight of 1 to class 0 and a weight of $\frac{sum\  negative}{ sum\  positive}$ for class 1. For the Catboost algorithm we will also use the \emph{auto class weigth} with the \emph{sqrtbalanced} \eqref{eq:sqrtbalanced}  for the values that multiply the objective function \cite{prokhorenkova2018catboost}.
\begin{equation}
CW_{k}  =  \sqrt{\frac{max^{k} _{c=1} (\sum_{t_i =c} {w_i})} {\sum_{t_i =k} {w_i}}}
\label{eq:sqrtbalanced}
\end{equation}

All the methods will be combined with all the models to find the best performance. 

The second problem that arises when dealing with unbalanced datasets is the metrics used to evaluate model performance. While most algorithms maximize accuracy, this metric can not be used in unbalanced datasets as it will build a dummy classifier that will only predict the majority class. Instead,  the interest is not only to make the largest amount of good classifications overall but have a good proportion of well-classified samples for both the majority and minority class. Hence, the selection of the metrics will be key to evaluate the results. This topic will be widely covered in section five.

\section{Model}

Prediction models seek to find a relation linking the target variable with the independent ones.
Once we obtain an estimate of such dependence, we may use it to predict the value of the target variable for new individuals.
\par

Credit Risk models consider the borrower's default as target variable (1 if the default occurred, 0 otherwise). Generally, the models try to predict the probability of default (PD), which can assume any continuous value from 0 to 1.
\medskip

This section provides a theoretical background for the models employed in this paper.
\par

For the objective of this analysis, binary classification is preferable. For this reason, a threshold between 0 and 1 should be defined, generally 0.5, which will mark the division of the two classes.

We will divide the dataset and use $75\%$ of it for training and the remaining $25\%$ for testing. To obtain more accurate results the model is trained using a 5-fold cross-validation comparison of the different techniques and resampling methods. 

The algorithms have been implemented using the Scikit-learn library and the Tensorflow-Keras machine learning software frameworks. 

\subsection{Logistic Regression}

One of the most common techniques in Credit Risk is Logistic Regression. We provide a brief model's overview, while the interested reader may refer to \cite{visani2019explanations} for an in depth analysis of the Logistic Regression, applied on Credit Scoring data.

The key idea is to model the conditional mean $E(y|x)$, from now on $\pi(x)$, wrt the independent variables $X$. 

To do so, we shall consider some constraints on $\pi(x)$, namely it is bounded in the interval $[0,1]$.
Therefore, we employ the transformation $ logit(x) = log \left( \frac{\pi(x)}{1-\pi(x)} \right)$ which takes values in $[-\infty,+\infty]$. It is now legit to assume a linear relationship between $ logit(x)$ and the independent variables $X$, namely $log \left( \frac{\pi(x)}{1-\pi(x)} \right) = X\beta$.
The formula may be rewritten with respect to $\pi(x)$, which higlights the non linear relationship with the $X$ variables, namely $\pi(x_i)= E(y|x) = \frac{e^{X\beta}}{1+ e^{X\beta}}$.

The vector $\beta$ contains the intercept and the coefficients for each $X$ variable, which represent the slope of the non linear relationship. We should find the best values for the $\beta$ parameters, this is usually done by the Newton-Rhapson optimization framework, which finds the parameters achieving the maximum log-likelihood for our dataset \cite{greene2008econometric}.
\medskip

While Logistic Regression made a good work on associating a probability to a prediction and it didn't assume $Y$ was a linear combination of $X$, it has the limitations of not being able to capture non-linear trends as it considers that there is linearity between the independent variables and log odds \cite{bolton2010logistic}. 
As a consequence, we present a range of ML models intending to overcome this major drawback.
\medskip

On a completely different line, the following models attempt to recover the relation between $\pi(x)$ and the $X$ without assuming it to belong to a class of parametric functions. In this way, they are non-parametric and generically called Machine Learning models.

\subsection{Random Forest}

The Random Forest exploits Decision Tree models - very simple non-linear models which cut the geometrical space of the $X$ variables recursively, with the aim to cluster together regions with the same value of the target variable -.\\
Random Forest aggregates together many simple decision trees, using the Bagging procedure, to increase their prediction power. In addition, the trees are generated more arbitrarily, choosing randomly the split variable at each node. This procedure increases the diversity among trees and consequently improves the performance of the ensemble model. \\
The model is capable to represent highly non-linear functions and usually achieves good predictions. Another strong point of Random Forest is overfitting: thanks to the bagging procedure, the model does not suffer a decrease in accuracy when expanding the number of trees.

In order to improve the performance of the model, the exploratory analysis will be done. Some of the important parameters to be tested are the number of estimators and their depth. Hence, it will be crucial finding the right amount of decision trees and leaves to prevent overfitting.

\subsection{Gradient Boosting}

On the same page, also Gradient Boosting employs Decision Trees. Differently from Random Forest, the model utilises the boosting procedure as aggregation technique: small trees are sequentially added to the model to reduce the loss, while keeping the previous trees fixed. Each tree focuses more on the individuals which have been badly predicted from the previous trees.
This training phase is guided by the Gradient on the errors of the preceding trees, hence the name Gradient Boosting. For binary classification problems, Gradient Boosting uses the Cross-Entropy Loss (\ref{eq:log loss}) as loss function:
\begin{equation}
L_{CE} (p,y) =  {\sum_i {y_i }{log(p_i)}}
\label{eq:log loss}
\end{equation}
The technique usually achieves very good results, although it is prone to overfitting. To control it, it is good practice to use early-stopping on the number of trees, during the training phase. Grid Search is suggested to tune the other hyperparameters \cite{friedman2001greedy}. 
\medskip

For a more efficient implementation of the algorithm, the Light Gradient Boosting model will also be tested \cite{ke2017lightgbm}.
\medskip

\subsection{CatBoost}

CatBoost is an implementation of Gradient Boosting machine learning tool developed by the Russian tech company Yandex. Different incentives lead to the creation of the algorithm. For instance: the treatment of data coming from various sources and the need for handling categorical variables. Moreover, it is constant to parameter changes and eliminates the costly task of tuning, showing great results from the first runs \cite{dorogush2018catboost}.

CatBoost uses oblivious decision trees \cite{kohavi1994bottom}. Compared to classic Decision Trees, the oblivious version imposes that nodes at the same height in the tree should use the same variable for the splitting. The modification is justified as to prevent overfitting, making it more stable to parameter changes \cite{kohavi1995oblivious}. Oblivious trees are an easily parallelizable algorithm, which allows training using GPUs, reducing the time to obtain a properly tuned model. 

CatBoost uses category-based statistics to handle categorical values. It considers that the encoding of the categorical features is better performed by the algorithm itself, rather than by humans. To do so,  it creates numerical features from the categorical ones, by using the category's number of occurrences in the dataset. 

The difference between the classical Gradient Boosting and CatBoost algorithm is that, in the former, the leaf values are calculated averaging the gradients in the current leaf which means it is an estimate of the gradient for all possible individuals in the leaf. This intrinsically introduces a bias, due to considering the model predictions on the same individuals using for training. To overcome the overfitting problem, CatBoost computes the gradient for each individual separately \cite{prokhorenkova2018catboost}. Then, the gradient will be based only on the individuals in the tree before the one being assessed. In practice, it trains the logarithm of the $n$ models, which are trained at the same time. Consequently, this approach will work well with small datasets as it is computationally expensive  \cite{dorogush2018catboost}.

\subsection{Neural Network}
Neural Networks are processing algorithms with an architecture that follows the brain biological structure. They are inspired to mimic the function of the human brain by feeding information through different layers and nodes. The simplest form is called multilayer perceptron (MLP) and represents a feed-forward network, namely the information flows, in a single direction and only once, through the framework. Its basic structure consists of an input layer followed by an undefined number of hidden layers and a final output layer that outputs the predicted value of the dependent variable. Each layer possesses different nodes which are responsible for computing a weighted sum of the input information received from the nodes of the foregoing layers. This result will be sent to a non-linear activation function. The process is repeated for all the layers, until the output layer.

More advanced networks change the propagation scheme through the network, breaking the feed-forward mechanism and allowing for more complex interactions among the nodes.

Back-propagation is the most employed neural training method, it consists of an efficient propagation of the gradient-based errors through the nodes in a backward fashion, which allows for optimization of the network's weights. Different types of gradient descent can be used, in our implementation, we employ the Stochastic Gradient Descent (SGD).

\section{Results}
 
To perform the evaluation of the different models we propose the AUC-ROC curve and the GINI index.
We present a comparison of the four methods previously described. 

\subsection{Evaluation Metrics}

To quantify the number of correct classifications the model makes for each class, counting the overall number is not enough. For instance, if the accuracy is high but all the correctly classified samples come from the majority class, the model is useless and does not provide any relevant output for the task. 
Therefore, we will consider different metrics that recognise the relevance of the model predictions. 

The confusion matrix aggregates all the classification information in a table. Rows represent the true values and columns the predicted ones. Each element in the table denotes a different option. True positives and negative will be the individuals correctly classified by the model as good and bad payers respectively. On the contrary, false positive and negative will be good or bad customers who were incorrectly identified.

However, the confusion matrix can't help on evaluating the performance.

Consequently, the metrics proposed are the ROC-AUC and Gini index. The ROC is a probability curve in the form of a graph that represents the proportion of the true positive rate $tpr = {\frac{TP}{TP+FN} }$ against the false positive rate $fpr =  {\frac{FP}{FP+TN} }$. It shows from all the possible thresholds the performance of the model making the final result invariant to it. 
To compute the model efficiency for separating the classes, the AUC (Area Under the Curve) formula evaluates the aggregated response of the model, given all the thresholds from 0.5 to 1, to classify a trustworthy customer over an untrustworthy one.

The Gini index \eqref{eq:gini} is derived from the AUC and it is a standard metric used in risk assessment. As AUC, it gives the ratio between true positive and true negatives but in the range between 0 and 1, making it more intuitive.
\begin{equation}
GINI =   2 \cdot AUC-1
\label{eq:gini}
\end{equation}

\subsection{Experiments}
\begin{figure}[h]
    \centering
    \includegraphics[width=80mm,scale=0.5 ]{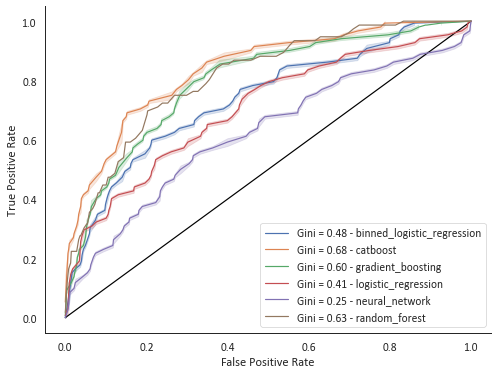}
    \caption{The illustration shows the AUC-ROC curve with the true positive against the false positive rate for the different models.}
    \label{fig:EXPERIMENT 1}
    
\end{figure}

The first experiment runs the 4 models. The classical Logistic Regression is compared with Gradient Boosting, CatBoost and the Neural Network models. 
{\bf Figure \ref{fig:EXPERIMENT 1}} shows how Gradient Boosting is the best in class, with CatBoost and Neural Networks following close while Logistic Regression has the worst performance in terms of AUC. 

\begin{table*}[ht]

\centering
\begin{tabular}{p{0.23\linewidth}p{0.23\linewidth}p{0.15\linewidth}p{0.15\linewidth}}
\hline
\\[-1em]
 \bf Classifiers & \bf Resampling &\bf GINI  (std) &\bf 20F GINI  (std) \\ 
 \\[-1em]
\hline\hline
\\[-1em]
 Logistic Regression  & No Resampling & 0.41 (0.11) & 0.47 (0.11)  \\
 & BorderlineSMOTE & 0.36 (0.10) & 0.34 (0.13)\\
 \\[-1em]
& SVMSMOTE & 0.33 (0.18) & 0.35 (0.15) \\
\\[-1em]
\hline
\\[-1em]
 Logistic Regression (Binned)  & No Resampling & 0.44 (0.11) & \\
 & BorderlineSMOTE & 0.48 (0.15) & \\
 \\[-1em]
& SVMSMOTE & 0.45 (0.08) &  \\
\\[-1em]
\hline\\[-1em]
Random Forest  & No Resampling & 0.62 (0.12) & 0.62 (0.14)  \\
& BorderlineSMOTE & 0.61  (0.08) & 0.57  (0.13)\\
\\[-1em]
& SVMSMOTE & 0.54  (0.12)& 0.62  (0.16)\\
\\[-1em]
& Oversampling Minority & 0.63  (0.10)  & 0.58  (0.17)\\
\\[-1em]
& Undersampling Majority & 0.59  (0.16) & 0.63  (0.13) \\
\\[-1em]
\hline
\\[-1em]
Gradient Boosting  & No Resampling & 0.58 (0.11) & 0.62 (0.13)  \\
& BorderlineSMOTE & 0.56  (0.07) & 0.55  (0.13)\\
\\[-1em]
& SVMSMOTE & 0.52  (0.10) & 0.60  (0.14)\\
\\[-1em]
& Oversampling Minority & 0.60  (0.10) & 0.53  (0.22) \\
\\[-1em]
& Undersampling Majority & 0.57  (0.09)& 0.58  (0.09)\\
\\[-1em]
\hline
\\[-1em]
CatBoost  & No Resampling & 0.68 (0.08) & 0.66 (0.10)  \\
\\[-1em]
& Class-weight & 0.56  (0.10) & 0.58  (0.14)\\
\\[-1em]
& BorderlineSMOTE & 0.60  (0.03) & 0.59  (0.13) \\
\\[-1em]
& SVMSMOTE & 0.57  (0.12) & 0.62  (0.12)\\
\\[-1em]
& Oversampling Minority & 0.64  (0.07) & 0.55  (0.14)\\
\\[-1em]
& Undersampling Majority & 0.59  (0.16) & 0.60  (0.13)\\
\\[-1em]
\hline
\\[-1em]
Neural Networks  & Class weight & 0.20  (0.20)& 0.52  (0.11)\\
\\[-1em]
\hline
\end{tabular}

\caption{Comparative table of different resampling and classification methods with their corresponding Gini index and error. The third column refers to the models trained on the dataset after the pre-processing step, with a total of 79 features. The fourth column shows the GINI for the model trained on the 20 most important features, obtained through the Shapley values built on the Catboost model with all the features. The error estimated as the standard deviation from the results of the 5 fold cross-validation is detailed in parentheses.}
\label{tab:GINI comparison}
\end{table*}

The third column in {\bf Table \ref{tab:GINI comparison}} reveals that the highest GINI was obtained using standard CatBoost with no resampling
Of the four initial classifiers, in all cases, BorderlineSMOTE resampling method obtains better performance than its counterpart SVMSMOTE.

\begin{figure}[h]
    \centering
    \includegraphics[width=80mm,scale=0.5 ]{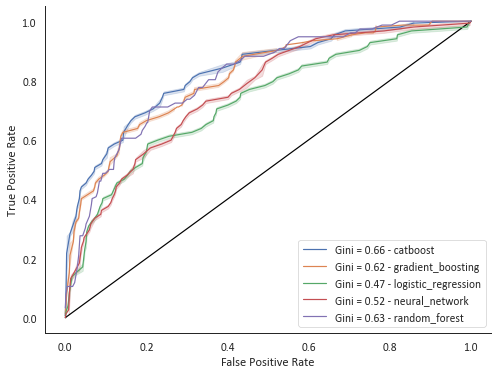}
    \caption{The illustration shows for the 20 most important features the AUC-ROC curve with the true-positive rate against the false-positive rate for the different models.}
    \label{fig:EXPERIMENT 2}
\end{figure}
According to the explainable AI model pursued in the paper, a second experiment will compute the results using only 20 features selected using the feature importance calculated with the SHAP values built on the Catboost model. {\bf Figure \ref{fig:EXPERIMENT 2}} presents the results after the dimensionality reduction. When comparing to the initial approach, the majority of models surprisingly maintained or insignificantly reduced its accuracy while introducing benefits, especially for the interpretation and training time.

Looking at the fourth column of {\bf Table \ref{tab:GINI comparison}},  standard CatBoost method again outperforms the others, followed by Gradient Boosting and Random Forest. With a few exceptions, our results denote that decreasing the number of features did not affect consistently the model's performance. Reducing the number of features decreases the gap between the different models, with or without pairing them with a resampling technique.

\section{Explainability}

Black box machine learning models have the major disadvantage of not being able to explain the rationale behind a prediction. The core intention in machine learning is about providing the best possible forecast, but in many real-life scenarios, there is also a need for useful information about the decision \cite{lipton2018mythos}.

Explainability can be defined as giving human-understandable motivations of how given attributes of an individual are related to its model prediction \cite{10.1145/3236009}. While interpretability stands for providing some meaning to the human in a way it can be understood, explainability goes a step further by finding a human-comprehensive way to understand the decision-maker algorithm \cite{arrieta2020explainable}.
It can be distinguished between local and global. While the former explains the reasons for a specific decision on a single individual, the latter focuses on providing some meaning for the whole model's logic to grasp the grand scheme of the algorithm \cite{8466590}.

In the Credit Scoring field, this topic is relatively new and particularly important, given the strict regulation on the topic especially in the European community such as European General Data
Protection Regulation (GDPR) and Ethics guidelines for trustworthy AI \cite{Ethical}. 
In addition to the regulatory issues, banks and financial institutions take into high consideration the chance of understanding the model reasoning: it allows to provide data-driven insights comprehensible by the human operators.

The explainability topic in Credit Scoring has been already tackled in previous works, using in particular the LIME framework \cite{ribeiro2016should} and its new extensions guaranteeing more stability for the explanations \cite{visani2020statistical} \cite{visani2020optilime}. In this contribution, we propose a different line exploiting the SHAP algorithm: a tool assigning the importance level to each feature in the model, namely how much the variable contributes in achieving a good prediction.

SHAP (SHapley Additive exPlanations) \cite{ NIPS2017_7062} is based on game theory, in particular on the Shapley values \cite{shapley1953value}. The original framework was developed to redistribute the gain of a cooperative game among players, in a rightful way. The same idea is borrowed in SHAP, where the aim is to decompose the prediction of the ML model among the features involved. 
\begin{equation}\label{eq:additive features}
f(x) =  \phi_0 + \sum_{i=1}^M \phi_i
\end{equation}

In {\bf Equation \eqref{eq:additive features}}, $\phi_0$ represents the baseline while $\phi_i$ is the specific contribution of the feature $i$ to produce the ML prediction $f(x)$ for the single individual whose feature vector is represented by $x$. \\
Regarding the $\phi_i$ calculation, SHAP exploits the Shapley algorithm:
\begin{equation}
\resizebox{.8\hsize}{!}{$\phi_i=\sum_{S\subseteq F\setminus \{i\} } {\frac{|S|!(|F|-|S|-1)!}{|F|!}}  {[{f_{S \cup \{i\}}(x_{S \cup \{i\}})} - {f_S (x_S)}]} $}
\label{eq:shapley values}
\end{equation}
The idea is to consider an ML model with a restricted set of variables $S$ out of $F$ (complete set), not containing the $i$-th feature. We evaluate the difference in prediction between the model using only the $S$ variables $f_S(x)$ and the same model adding also the $i$-th feature $f_{S \cup \{i\}}$. The difference $f_{S \cup \{i\}}(x) - f_S(x)$ is attributed to the $i$-th variable. \\
Ideally we should consider any possible set of features $S$ and average each difference in prediction caused by the $i$-th variable. This is not feasable from a computational point of view, therefore SHAP performs a random sampling of the possible sets $S$ and computes the average on their prediction difference. Hence, we obtain an estimate of the $i$-th variable importance.
\medskip

This procedure works for any kind of ML model, at the cost of an elevated computation time. \\
A recent improvement of the SHAP technique consists in the TreeSHAP algorithm \cite{lundberg2018consistent}, which provides exact calculation of the feature importances for Tree-based models. The key intuition is to exploit the tree structure, to calculate the importance of all the possible sets $S$, in just a single pass. Along with the exact computation advantage, also the running time is drastically reduced.
\medskip

SHAP is a local technique, since it decomposes a single prediction at a time. Although, there are ways to generalize the individual feature importances to the entire dataset, namely calculating the average importance of the variable among all the individuals. In doing so, we obtain a global feature importance, which quantifies the relevance of each variable for the ML model.
\medskip

In this paper, the SHAP technique will be used to solve two different research questions: firstly we use SHAP global feature importance to rank the dataset variables in order to make feature reduction and keep only the 20 most relevant ones, secondly SHAP on single individuals is going to provide insights on the rationale of the final ML model.
\medskip

From the results obtained from the comparison of the different models, CatBoost was selected for the interpretable machine learning approach with TreeSHAP.
The model is fitted with the training data, leaving the test for SHAP predictions to establish the additive features.

\subsection{Global Interpretation}
SHAP provides a set of useful plots for local and global interpretations of the feature's contribution to the model. 

\begin{figure}[h]
    \centering
    \includegraphics[ width=80mm ]{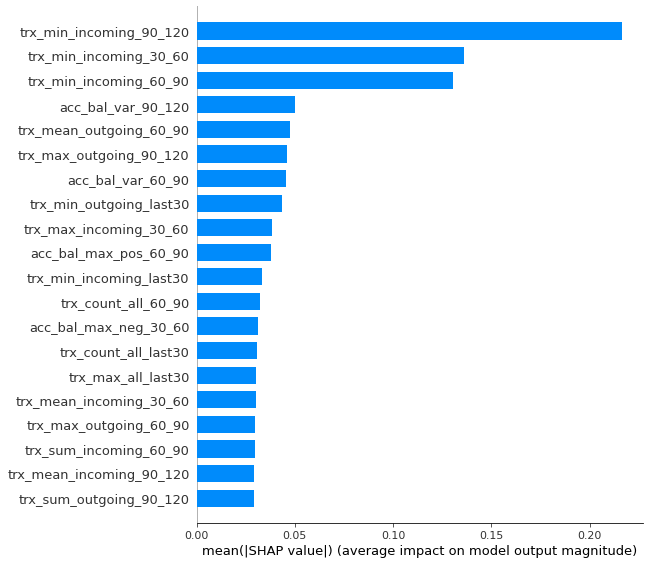}
    \caption{The illustration shows the importance of each feature to the output magnitude. }
    \label{fig:shap summary plot}
\end{figure}
The summary plot on { \bf Figure \ref{fig:shap summary plot}} gives the ranking for the absolute average feature's impact on the model output. Of the 20 most important features, the incoming transactions with the lowest amount, i.e., the incomes of the subject are low, carry the largest information. Closely following, the variation of the account balances. 
\begin{figure}[h]
    \centering
    \includegraphics[width= 80mm ]{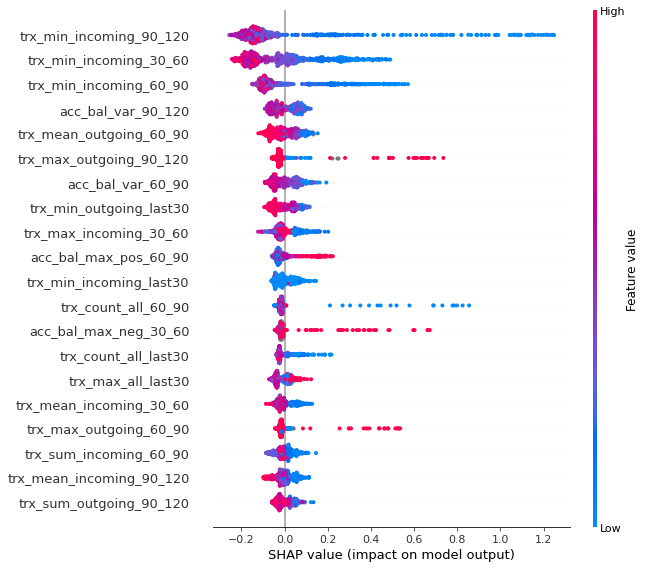}
    \caption{SHAP values for each feature (each dot represents the value for a single individual)}
    \label{fig:shap impact features}
\end{figure}

{\bf Figure \ref{fig:shap impact features}} summarizes the Shap values for every feature for the whole dataset. The dots shown in each variable's row represent the individual subjects in the dataset. On the $Y$ axis the features are ordered by the mean absolute SHAP value.
The horizontal $X$ axis collect the SHAP values reflecting the positive or negative impact of each feature on model prediction. A positive SHAP value increases the probability of the sample to be considered a bad customer while the opposite goes for the negative. Furthermore, a longer tail reflects a higher impact on the prediction.
The colormap bar moves from blue to red as the feature value increases. High feature values on the positive side of the $X$ axis have a positive correlation with the dependent variable $Y$. Hence, there is a positive effect of that values to the prediction, promoting a bad account. For instance, in the feature \textit{acc\_bal\_max\_neg\_30\_60} (appendix \ref{appendix:graph}), the higher the value of its samples, the higher the SHAP values and therefore, higher the probability of a bad individual. 
On the contrary, negative correlations push the instances to be considered as good customers. The $1^{st}$ variables in the ranking (\textit{trx\_min\_incoming\_90\_120}) are an example of this negative influence. The higher the Shapley values for these variables, the higher the force they have to consider an account a good one.

\begin{figure}[h]
    \centering
    \includegraphics[width= 80mm]{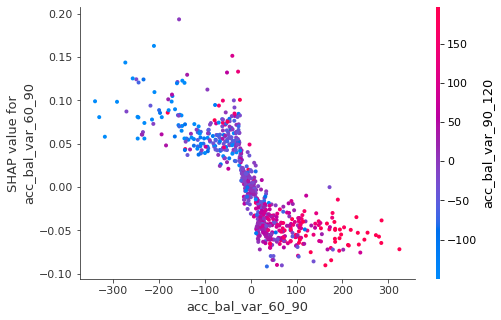}

    \caption{Scatter plot of the Shap values of the variation of the account balance between 60 and 90 days before the loan was granted against the the same variation in the previous 30 days. }
    \label{fig:dependence max AB}
\end{figure}

Dependence plots are a tool that helps understand the global allocation of shap values to a particular account given the general behaviour. In {\bf Figure \ref{fig:dependence max AB}} we show the relation between the variation of the account balance (VAB) values and their Shapley values. As expected, the model assigns negative SHAP values, influencing the account to be a good customer when the VAB is higher than zero. Looking at the variation of the previous window of time (between 90 and 120 days), for an account with relatively high values (pink and purple samples), the SHAP values are negative or close to zero and predict a good customer and the trend seems to remains constant.

The selection of the 20 most important features can bring some knowledge about the characteristics the model found to be more useful for clustering the accounts. Overall, all the three windows available were selected. Therefore, having the data split into shorter periods brings relevant information, since the whole history of the subject can help discriminate among good and bad payers. 
A second remark refers to the type of feature. We can distinguish 3 different categories: variables referring to the transactions with their sign, to the transaction balance or the account balance variation. Of the 20 variables, 13 were associated with the transactions with their sign, 4 to the account balance and only 3 to the transaction balance. 
In response to the type of operation performed to obtain the feature \emph{(min, max, mean)},  most significant features were related with operations looking for minimums, followed by the mean. Hence, we can interpret that the negatives/positives and minimum values were highly more relevant for the model to extract meaningful information.

\subsection{Local Interpretation}
In this subsection, the four different types of possible accounts of the confusion matrix \emph{(TP, TN, FP, FN)} are discussed. 
\subsubsection{True Positive}
The \emph{True Positive} is an account of a bad customer that was correctly identified by the model as being bad.
A random account from the test dataset has been chosen to exemplify the behaviour of the model showing a 0.57 probability to be considered bad.  {\bf Figure \ref{fig:bad was bad}} shows that the feature with the highest impact for predicting a bad customer was \textit{trx\_min\_incoming\_90\_120} (appendix \appendix{}). In the second position, \textit{trx\_max\_outgoing\_60\_90} reflects the importance not only of the incomes, but also on the expenses of the account. 
On the other hand, given that the \textit{trx\_sum\_outgoing\_90\_120} is high compared to the following periods (total sum of outgoing transactions between 90 and 120 days before the loan application date), this feature impacts the model to consider the account a good one. However, as the influence is not enough, the features on the left have more power and succeed in the prediction task.
\begin{figure}[h]
    \centering
    \includegraphics[width= 80mm]{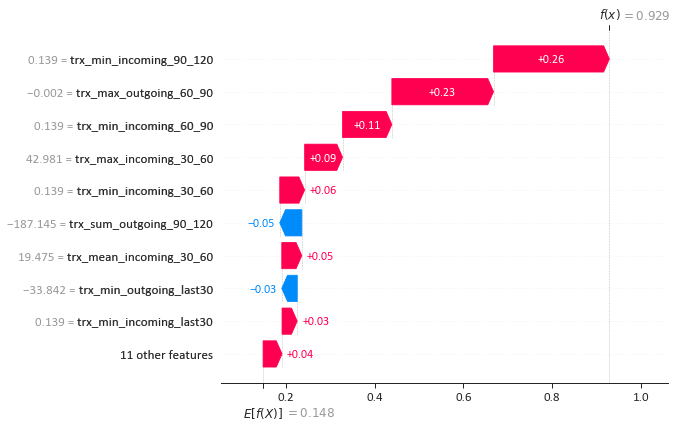}

    \caption{\emph{True Positive}. Waterfall plot of a bad account categorized as bad by the model. }
    \label{fig:bad was bad}
\end{figure}

\begin{figure}[h]
    \centering
    \includegraphics[width= 80mm]{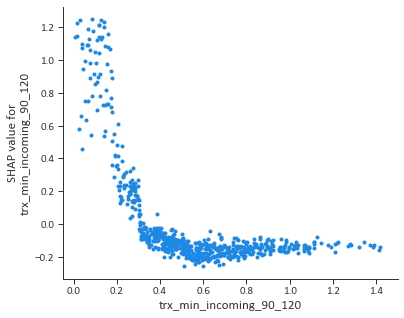}
    \caption{Dependence plot of the \textit{trx\_min\_incoming\_90\_120}, minimum value of the incoming transactions between 90 and 120 days before the loan application date. }
    \label{fig:dependence most imp feature true bad}
\end{figure}

To explain the importance of the first feature on {\bf Figure \ref{fig:bad was bad}}, we will take advantage of the global perspective that a dependence plot offers.
When features are strongly correlated we have to be careful about the interpretation of the feature importance.  Considering that the interaction with other features increases the feature's value, in {\bf Figure \ref{fig:dependence most imp feature true bad}} we analysed how the SHAP and feature values relate in the circumstances where the interaction with other features are considered.

A general look at the graph shows a pick at 0, meaning that the model gives importance to this feature when it finds values of the \textit{trx\_min\_incoming\_90\_120} equal or close to zero. Probably, this is an indicator of a not very active account or one that doesn't move large quantities, marking it as less trustful. When this feature moves away from zero, it becomes less important for the model to use it for its predictions. Interestingly, this feature becomes irrelevant when the values differ from the ones explained. In this cases, the contribution to the prediction is usually negative, pushing the model towards a good customer. 

\subsubsection{True Negative}
\begin{figure}[h]
    \centering
    \includegraphics[width= 80mm]{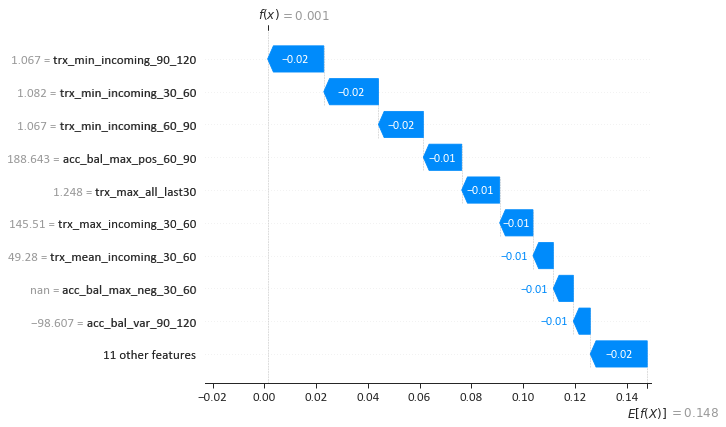}
    \caption{\emph{True Negative}. Waterfall plot of a good account categorized as good by the model. }
    \label{fig:good as good}
\end{figure}

The \emph{True Negative} is an account of a good customer that was correctly identified by the model. 
The account on {\bf Figure \ref{fig:good as good} } shows that the minumum of the incoming transactions is always much greater than zero in all the three reference periods. Those minima, together with the absence of negative account balance (\textit{acc\_bal\_max\_neg\_30\_60}), are indicators that the customer is likely to be good with only a  0.001 probability of misbehaviour. 

\subsubsection{False Positive}
\begin{figure}[h]
    \centering
    \includegraphics[width= 80mm]{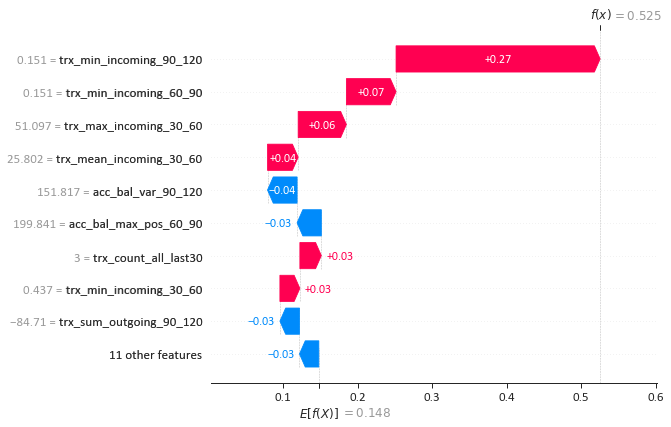}

    \caption{\emph{False Positive}. Waterfall plot of a good account categorized as bad by the model. }
    \label{fig:bad was good}
\end{figure}

The \emph{False Positive} is an account of a good customer that was incorrectly identified as bad. The account in {\bf Figure \ref{fig:bad was good}} is an interesting example of this type as it was wrongly classified with a predicted probability of 0.525. Some signs of this failed attempt could be the fact that the minimum of the incoming transaction amounts are too close to zero and that the account was inactive, reporting a low number of transactions in the period immediately before the loan application date (\textit{trx\_count\_all\_last30}). Therefore, the model choice was in correspondence with the numbers, even though the customer behaved in an unpredicted way given its transactions trends. 

\subsubsection{False Negative}
\begin{figure}[h]
    \centering
    \includegraphics[width= 80mm]{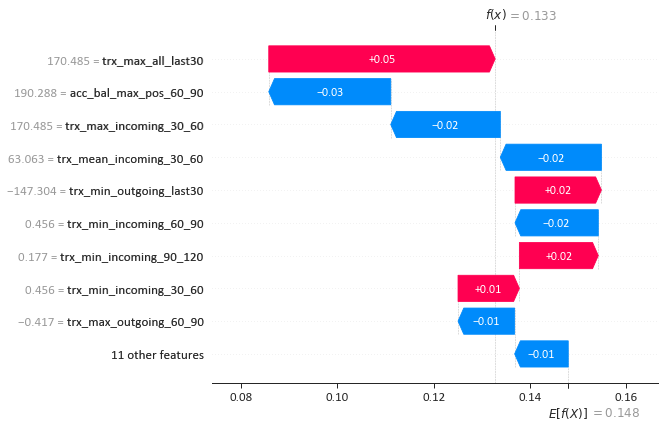}

    \caption{\emph{False Negative}. Waterfall plot of a bad account categorized as good by the model. }
    \label{fig:good was bad}
\end{figure}

The \emph{False Negative} is an account of a bad customer that was incorrectly identified as good. In {\bf Figure \ref{fig:good was bad}} this client type is exemplified with \textit{trx\_max\_all\_last30} (appendix \ref{appendix:graph}) being the feature with a major impact toward what would have been a correct guess. 
Nevertheless, the high value of \textit{trx\_max\_incoming\_last30}, i.e. the same value got the month before, has the opposite effect, despite showing the same amount. That is a clear example of a model choice that is not interpretable from the model's output. Hence, we reported the SHAP dependence plot of these 2 variables to acquire more information.

\begin{figure}[h]
    \centering
    \includegraphics[width= 80mm]{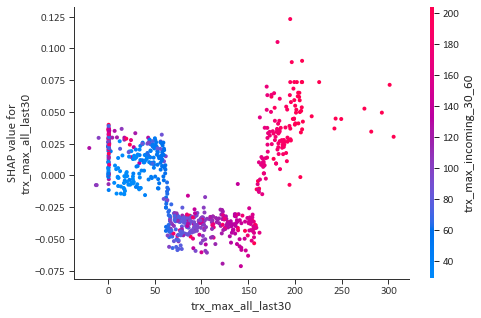}

    \caption{Dependence plot between the Shapley values of the maximum of the incoming (all contain both incoming and outgoing) transaction amount for the second month against the first month of a bad account categorized as good by the model.}
    \label{fig:dependence ABm ABf}
\end{figure}
{\bf Figure \ref{fig:dependence ABm ABf}} explains how the model makes a prediction when considering more than one feature. The colourmap on the horizontal axes shows the magnitude of the maximum transaction amount for all the features, i.e. for the incoming ones that have positive sign, in the first month.

Points on the left have a higher maximum incoming value in the second months and, as expected, are coloured in blue meaning that they also have a low value for the previous month. The same correlation is shown when we move towards the right side. 
Surprisingly, points with maximum amount near zero don't influence the model to make a decision, and therefore, the Shapley values are very close to zero. As we get closer to 75 \EUR, we observe a drop and a plateau with negative  Shapley values allocated to these accounts influencing the prediction towards a good client. However, this just pumps for the accounts where we find similar values in both months (with a colour moving from blue to purple). On the contrary, when one of the months turns purple, with extremely high incomes while we move to the right extreme of the axes, the behaviour changes and the model predicts a bad.  That is very interesting inside as we may expect high incomes values to be considered good and low bad, but the algorithm has learned that extreme incomes are normally not associated with good customers. In contrast, in the case of having maximum incoming transactions in the range 75 \EUR - 275 \EUR, it doesn't hesitate to consider it a good customer. Looking at the result from a business side, we may argue that customers with a standard income are more likely to get credit and be trusted by the bank. On the other hand, clients who show this extreme behaviour on the income side are less trustful.

\section{Conclusions}
In this project, a machine-learning-based method for credit scoring has been suggested. To overcome the legal requirements that obligate financial institutions to explain the basis of every rejected loan application, we presented an explainable model that can break down a prediction by showing the impact of each feature. 

Moreover, our research has highlighted the importance of choosing or not choosing the adequate resampling technique to solve the unbalance of the dataset. 
We have provided further evidence that regardless of the number of features used, boosted models outperform Linear Models, Decision Trees and Neural Networks. Despite some inconsistencies between the AUC comparisons, with cross-validation, we confirmed the outstanding performance of Catboost over its boosted family-algorithms.

Our experiments coincide with previous results defending that boosted models can be more accurate than Neural Networks at the same time of being more interpretable than Linear Models \cite{lundberg2020local2global}.
Our research on Neural Networks suggests that it should be not considered as the preferred model without notably increasing the size of the dataset.

The strength of our work lies in the explainability section for which we used SHAP to interpret the model predictions from a global and local perspective. The findings are not transferable to all credit scoring models because they provide an adjusted understanding of the outcomes for the selected bank. Consequently, it revealed the importance of not building a universal answer and de-mystifies the assumption of a unique solution. One promising application of our technique would be to understand the bank-customer relationship. Not only by understanding the behaviour of an account but also, the behaviour of clients as a collective that interacts with a financial entity. 




\begin{acknowledgments}
We acknowledge financial support by CRIF S.p.A and Universit\`a degli Studi di
Bologna. We would like to thank Robert Lesi and Annamaria Boni for their valuable advice and ongoing collaboration with the experimental work.
\end{acknowledgments}


\bibliography{biblio}

\begin{thebibliography}{42}
\providecommand{\natexlab}[1]{#1}
\providecommand{\url}[1]{\texttt{#1}}
\expandafter\ifx\csname urlstyle\endcsname\relax
  \providecommand{\doi}[1]{doi: #1}\else
  \providecommand{\doi}{doi: \begingroup \urlstyle{rm}\Url}\fi

\bibitem[Thomas et~al.(2017)Thomas, Crook, and Edelman]{thomas2017credit}
Lyn Thomas, Jonathan Crook, and David Edelman.
\newblock \emph{Credit scoring and its applications}.
\newblock SIAM, 2017.

\bibitem[Gup and Kolari(2005)]{gup2005commercial}
Benton~E Gup and James~W Kolari.
\newblock \emph{Commercial banking: The management of risk}.
\newblock John Wiley \& Sons Incorporated, 2005.

\bibitem[Barron and Staten(2003)]{barron2003value}
John~M Barron and Michael Staten.
\newblock The value of comprehensive credit reports: Lessons from the us
  experience.
\newblock \emph{Credit reporting systems and the international economy},
  8:\penalty0 273--310, 2003.

\bibitem[Gomber et~al.(2018)Gomber, Kauffman, Parker, and
  Weber]{gomber2018fintech}
Peter Gomber, Robert~J Kauffman, Chris Parker, and Bruce~W Weber.
\newblock On the fintech revolution: Interpreting the forces of innovation,
  disruption, and transformation in financial services.
\newblock \emph{Journal of Management Information Systems}, 35\penalty0
  (1):\penalty0 220--265, 2018.

\bibitem[Khandani et~al.(2010)Khandani, Kim, and Lo]{KHANDANI}
Amir~E. Khandani, Adlar~J. Kim, and Andrew~W. Lo.
\newblock Consumer credit-risk models via machine-learning algorithms.
\newblock \emph{Journal of Banking \& Finance}, 34\penalty0 (11):\penalty0 2767
  -- 2787, 2010.
\newblock ISSN 0378-4266.
\newblock \doi{https://doi.org/10.1016/j.jbankfin.2010.06.001}.

\bibitem[Bernthal et~al.(2005)Bernthal, Crockett, and Rose]{cclifestyle}
Matthew Bernthal, David Crockett, and Randall Rose.
\newblock Credit cards as lifestyle facilitators.
\newblock \emph{Journal of Consumer Research}, 32:\penalty0 130--145, 06 2005.
\newblock \doi{10.1086/429605}.

\bibitem[Wang et~al.(2015)Wang, Xu, and Zhou]{wang2015large}
Hong Wang, Qingsong Xu, and Lifeng Zhou.
\newblock Large unbalanced credit scoring using lasso-logistic regression
  ensemble.
\newblock \emph{PloS one}, 10\penalty0 (2):\penalty0 e0117844, 2015.

\bibitem[AI(2019)]{Ethical}
HLEG. AI.
\newblock Ethical guidelines for trustworthy ai.
\newblock \emph{European Commission}, Apr. 2019.

\bibitem[Kingston(2017)]{AIandLaw}
John Kingston.
\newblock Using artificial intelligence to support compliance with the general
  data protection regulation.
\newblock \emph{Artificial Intelligence and Law}, pages 1--15, Sep. 2017.
\newblock \doi{10.1007/s10506-017-9206-9}.

\bibitem[Polasik et~al.(2020)Polasik, Huterska, Iftikhar, and
  Mikula]{polasik2020impact}
Micha{\l} Polasik, Agnieszka Huterska, Rehan Iftikhar, and
  {\v{S}}t{\v{e}}p{\'a}n Mikula.
\newblock The impact of payment services directive 2 on the paytech sector
  development in europe.
\newblock \emph{Journal of Economic Behavior \& Organization}, 178, 2020.

\bibitem[Frost et~al.(2020)Frost, Gambacorta, Huang, Shin, and
  Zbinden]{10.1093/epolic/eiaa003}
Jon Frost, Leonardo Gambacorta, Yi~Huang, Hyun~Song Shin, and Pablo Zbinden.
\newblock {BigTech and the changing structure of financial intermediation}.
\newblock \emph{Economic Policy}, 34\penalty0 (100):\penalty0 761--799, 01
  2020.
\newblock ISSN 0266-4658.
\newblock \doi{10.1093/epolic/eiaa003}.

\bibitem[(IFC)(2019)]{CreditReporting}
International Finance~Corporation (IFC).
\newblock Credit reporting knowledge guide 2019.
\newblock \emph{World Bank Group}, 01 2019.

\bibitem[Huang et~al.(2018)Huang, Zhou, and Wang]{huang2018rfms}
Danyang Huang, Jing Zhou, and Hansheng Wang.
\newblock Rfms method for credit scoring based on bank card transaction data.
\newblock \emph{Statistica Sinica}, 28\penalty0 (4):\penalty0 2903--2919, 2018.

\bibitem[{Wang} et~al.(2019){Wang}, {Han}, {Liu}, and {Luo}]{LSTMarticle}
C.~{Wang}, D.~{Han}, Q.~{Liu}, and S.~{Luo}.
\newblock A deep learning approach for credit scoring of peer-to-peer lending
  using attention mechanism lstm.
\newblock \emph{IEEE Access}, 7:\penalty0 2161--2168, 2019.

\bibitem[Yang et~al.(2018)Yang, Zhang, et~al.]{yang2018comparison}
Shenghui Yang, Haomin Zhang, et~al.
\newblock Comparison of several data mining methods in credit card default
  prediction.
\newblock \emph{Intelligent Information Management}, 10\penalty0 (05):\penalty0
  115, 2018.

\bibitem[Petrocelli()]{financialdatawebaddress}
Liz Petrocelli.
\newblock 1999 czech financial dataset - real anonymized transaction.
\newblock URL
  \url{https://data.world/lpetrocelli/czech-financial-dataset-real-anonymized-transactions}.

\bibitem[Rosenberg and Gleit(1994)]{rosenberg1994quantitative}
Eric Rosenberg and Alan Gleit.
\newblock Quantitative methods in credit management: a survey.
\newblock \emph{Operations research}, 42\penalty0 (4):\penalty0 589--613, 1994.

\bibitem[Lundberg and Lee(2017{\natexlab{a}})]{lundberg2017unified}
Scott~M Lundberg and Su-In Lee.
\newblock A unified approach to interpreting model predictions.
\newblock In \emph{Advances in neural information processing systems}, pages
  4765--4774, 2017{\natexlab{a}}.

\bibitem[More(2016)]{more2016survey}
Ajinkya More.
\newblock Survey of resampling techniques for improving classification
  performance in unbalanced datasets.
\newblock \emph{arXiv preprint arXiv:1608.06048}, 2016.

\bibitem[Chawla et~al.(2002)Chawla, Bowyer, Hall, and
  Kegelmeyer]{chawla2002smote}
Nitesh~V Chawla, Kevin~W Bowyer, Lawrence~O Hall, and W~Philip Kegelmeyer.
\newblock Smote: synthetic minority over-sampling technique.
\newblock \emph{Journal of artificial intelligence research}, 16:\penalty0
  321--357, 2002.

\bibitem[Han et~al.(2005)Han, Wang, and Mao]{han2005borderline}
Hui Han, Wen-Yuan Wang, and Bing-Huan Mao.
\newblock Borderline-smote: a new over-sampling method in imbalanced data sets
  learning.
\newblock In \emph{International conference on intelligent computing}, pages
  878--887. Springer, 2005.

\bibitem[Nguyen et~al.(2011)Nguyen, Cooper, and Kamei]{nguyen2011borderline}
Hien~M Nguyen, Eric~W Cooper, and Katsuari Kamei.
\newblock Borderline over-sampling for imbalanced data classification.
\newblock \emph{International Journal of Knowledge Engineering and Soft Data
  Paradigms}, 3\penalty0 (1):\penalty0 4--21, 2011.

\bibitem[Prokhorenkova et~al.(2018)Prokhorenkova, Gusev, Vorobev, Dorogush, and
  Gulin]{prokhorenkova2018catboost}
Liudmila Prokhorenkova, Gleb Gusev, Aleksandr Vorobev, Anna~Veronika Dorogush,
  and Andrey Gulin.
\newblock Catboost: unbiased boosting with categorical features.
\newblock In \emph{Advances in neural information processing systems}, pages
  6638--6648, 2018.

\bibitem[Visani et~al.(2019)Visani, Chesani, Bagli, Capuzzo, and
  Poluzzi]{visani2019explanations}
Giorgio Visani, Federico Chesani, Enrico Bagli, Davide Capuzzo, and Alessandro
  Poluzzi.
\newblock Explanations of machine learning predictions: a mandatory step for
  its application to operational processes.
\newblock 2019.

\bibitem[Greene(2008)]{greene2008econometric}
William~H Greene.
\newblock The econometric approach to efficiency analysis.
\newblock \emph{The measurement of productive efficiency and productivity
  growth}, 1\penalty0 (1):\penalty0 92--250, 2008.

\bibitem[Bolton et~al.(2010)]{bolton2010logistic}
Christine Bolton et~al.
\newblock \emph{Logistic regression and its application in credit scoring}.
\newblock PhD thesis, University of Pretoria, 2010.

\bibitem[Friedman(2001)]{friedman2001greedy}
Jerome~H Friedman.
\newblock Greedy function approximation: a gradient boosting machine.
\newblock \emph{Annals of statistics}, pages 1189--1232, 2001.

\bibitem[Ke et~al.(2017)Ke, Meng, Finley, Wang, Chen, Ma, Ye, and
  Liu]{ke2017lightgbm}
Guolin Ke, Qi~Meng, Thomas Finley, Taifeng Wang, Wei Chen, Weidong Ma, Qiwei
  Ye, and Tie-Yan Liu.
\newblock Lightgbm: A highly efficient gradient boosting decision tree.
\newblock In \emph{Advances in neural information processing systems}, pages
  3146--3154, 2017.

\bibitem[Dorogush et~al.(2018)Dorogush, Ershov, and
  Gulin]{dorogush2018catboost}
Anna~Veronika Dorogush, Vasily Ershov, and Andrey Gulin.
\newblock Catboost: gradient boosting with categorical features support.
\newblock \emph{arXiv preprint arXiv:1810.11363}, 2018.

\bibitem[Kohavi(1994)]{kohavi1994bottom}
Ron Kohavi.
\newblock Bottom-up induction of oblivious read-once decision graphs.
\newblock In \emph{European Conference on Machine Learning}, pages 154--169.
  Springer, 1994.

\bibitem[Kohavi and Li(1995)]{kohavi1995oblivious}
Ron Kohavi and Chia-Hsin Li.
\newblock Oblivious decision trees, graphs, and top-down pruning.
\newblock In \emph{IJCAI}, pages 1071--1079. Citeseer, 1995.

\bibitem[Lipton(2018)]{lipton2018mythos}
Zachary~C Lipton.
\newblock The mythos of model interpretability.
\newblock \emph{Queue}, 16\penalty0 (3):\penalty0 31--57, 2018.

\bibitem[Guidotti et~al.(2018)Guidotti, Monreale, Ruggieri, Turini, Giannotti,
  and Pedreschi]{10.1145/3236009}
Riccardo Guidotti, Anna Monreale, Salvatore Ruggieri, Franco Turini, Fosca
  Giannotti, and Dino Pedreschi.
\newblock A survey of methods for explaining black box models.
\newblock \emph{ACM Comput. Surv.}, 51\penalty0 (5), August 2018.
\newblock ISSN 0360-0300.
\newblock \doi{10.1145/3236009}.

\bibitem[Arrieta et~al.(2020)Arrieta, D{\'\i}az-Rodr{\'\i}guez, Del~Ser,
  Bennetot, Tabik, Barbado, Garc{\'\i}a, Gil-L{\'o}pez, Molina, Benjamins,
  et~al.]{arrieta2020explainable}
Alejandro~Barredo Arrieta, Natalia D{\'\i}az-Rodr{\'\i}guez, Javier Del~Ser,
  Adrien Bennetot, Siham Tabik, Alberto Barbado, Salvador Garc{\'\i}a, Sergio
  Gil-L{\'o}pez, Daniel Molina, Richard Benjamins, et~al.
\newblock Explainable artificial intelligence (xai): Concepts, taxonomies,
  opportunities and challenges toward responsible ai.
\newblock \emph{Information Fusion}, 58:\penalty0 82--115, 2020.

\bibitem[{Adadi} and {Berrada}(2018)]{8466590}
A.~{Adadi} and M.~{Berrada}.
\newblock Peeking inside the black-box: A survey on explainable artificial
  intelligence (xai).
\newblock \emph{IEEE Access}, 6:\penalty0 52138--52160, 2018.

\bibitem[Ribeiro et~al.(2016)Ribeiro, Singh, and Guestrin]{ribeiro2016should}
Marco~Tulio Ribeiro, Sameer Singh, and Carlos Guestrin.
\newblock " why should i trust you?" explaining the predictions of any
  classifier.
\newblock In \emph{Proceedings of the 22nd ACM SIGKDD international conference
  on knowledge discovery and data mining}, pages 1135--1144, 2016.

\bibitem[Visani et~al.(2020{\natexlab{a}})Visani, Bagli, Chesani, Poluzzi, and
  Capuzzo]{visani2020statistical}
Giorgio Visani, Enrico Bagli, Federico Chesani, Alessandro Poluzzi, and Davide
  Capuzzo.
\newblock Statistical stability indices for lime: obtaining reliable
  explanations for machine learning models.
\newblock \emph{arXiv preprint arXiv:2001.11757}, 2020{\natexlab{a}}.

\bibitem[Visani et~al.(2020{\natexlab{b}})Visani, Bagli, and
  Chesani]{visani2020optilime}
Giorgio Visani, Enrico Bagli, and Federico Chesani.
\newblock Optilime: Optimized lime explanations for diagnostic computer
  algorithms.
\newblock \emph{arXiv preprint arXiv:2006.05714}, 2020{\natexlab{b}}.

\bibitem[Lundberg and Lee(2017{\natexlab{b}})]{NIPS2017_7062}
Scott~M Lundberg and Su-In Lee.
\newblock A unified approach to interpreting model predictions.
\newblock In I.~Guyon, U.~V. Luxburg, S.~Bengio, H.~Wallach, R.~Fergus,
  S.~Vishwanathan, and R.~Garnett, editors, \emph{Advances in Neural
  Information Processing Systems 30}, pages 4765--4774. Curran Associates,
  Inc., 2017{\natexlab{b}}.

\bibitem[Shapley(1953)]{shapley1953value}
Lloyd~S Shapley.
\newblock A value for n-person games.
\newblock \emph{Contributions to the Theory of Games}, 2\penalty0
  (28):\penalty0 307--317, 1953.

\bibitem[Lundberg et~al.(2018)Lundberg, Erion, and Lee]{lundberg2018consistent}
Scott~M Lundberg, Gabriel~G Erion, and Su-In Lee.
\newblock Consistent individualized feature attribution for tree ensembles.
\newblock \emph{arXiv preprint arXiv:1802.03888}, 2018.

\bibitem[Lundberg et~al.(2020)Lundberg, Erion, Chen, DeGrave, Prutkin, Nair,
  Katz, Himmelfarb, Bansal, and Lee]{lundberg2020local2global}
Scott~M. Lundberg, Gabriel Erion, Hugh Chen, Alex DeGrave, Jordan~M. Prutkin,
  Bala Nair, Ronit Katz, Jonathan Himmelfarb, Nisha Bansal, and Su-In Lee.
\newblock From local explanations to global understanding with explainable ai
  for trees.
\newblock \emph{Nature Machine Intelligence}, 2\penalty0 (1):\penalty0
  2522--5839, 2020.

\end{thebibliography}

\section{KPIs description}
\begin{table*}[ht]

    \centering
    \begin{tabular}{|p{0.3\linewidth}|p{0.7\linewidth}|}
\hline
\\[-1em]
{ \bf Code }  & {\bf Definition}\\
\\[-1em]

\hline
\\[-1em]
\textit{trx\_min\_incoming\_90\_120}& minimum incoming amount between 90 and 120 days before the application date.\\
\\[-1em]
\hline

\\[-1em]
\textit{trx\_min\_incoming\_30\_60}& minimum incoming amount between 30 and 60 days before the application date.\\
\\[-1em]
\hline

\\[-1em]
\textit{trx\_min\_incoming\_60\_90}& minimum incoming amount between 60 and 90 days before the application date.\\
\\[-1em]
\hline

\\[-1em]
\textit{acc\_bal\_var\_90\_120}& variation of the account balance between 90 and 120 days before the application date.\\
\\[-1em]
\hline

\\[-1em]
\textit{trx\_mean\_outgoing\_60\_90}& average outgoing amount between 60 and 90 days before the application date.\\
\\[-1em]
\hline

\\[-1em]
\textit{trx\_max\_outgoing\_90\_120}& maximum outgoing amount between 90 and 120 days before the application date.\\
\\[-1em]
\hline

\\[-1em]
\textit{acc\_bal\_var\_60\_90}& variation of the account balance between 60 and 90 days before the application date.\\
\\[-1em]
\hline

\\[-1em]
\textit{trx\_min\_outgoing\_last30}& minimum outgoing amount in the last 30 days before the application date.\\
\\[-1em]
\hline

\\[-1em]
\textit{trx\_max\_incoming\_30\_60}& maximum incoming amount between 30 and 60 days before the application date.\\
\\[-1em]
\hline

\\[-1em]
\textit{acc\_bal\_max\_pos\_60\_90}& maximum positive account balance between 60 and 90 days before the application date.\\
\\[-1em]
\hline

\\[-1em]
\textit{trx\_min\_incoming\_last30}& minimum incoming amount in the last 30 days before the application date.\\
\\[-1em]
\hline

\\[-1em]
\textit{trx\_count\_all\_60\_90}& number of transactions between 60 and 90 days before the application date.\\
\\[-1em]
\hline

\\[-1em]
\textit{acc\_bal\_max\_neg\_30\_60}& maximum negative account balance between 30 and 60 days before the application date.\\
\\[-1em]
\hline

\\[-1em]
\textit{trx\_count\_all\_last30}& number of transactions in the last 30 days before the application date.\\
\\[-1em]
\hline

\\[-1em]
\textit{trx\_max\_all\_last30}& maximum amount in the last 30 days before the application date.\\
\\[-1em]
\hline

\\[-1em]
\textit{trx\_mean\_incoming\_30\_60}& average incoming amount between 30 and 60 days before the application date.\\
\\[-1em]
\hline

\\[-1em]
\textit{trx\_max\_outgoing\_60\_90}& maximum outgoing amount between 60 and 90 days before the application date.\\
\\[-1em]
\hline

\\[-1em]
\textit{trx\_sum\_incoming\_60\_90}& sum of the incoming amounts between 60 and 90 days before the application date.\\
\\[-1em]
\hline

\hline
\\[-1em]
\textit{trx\_mean\_incoming\_90\_120}& average incoming amount between 90 and 120 days before the application date.\\
\\[-1em]
\hline

\hline
\\[-1em]
\textit{trx\_sum\_incoming\_90\_120}& sum of the incoming amounts between 90 and 120 days before the application date.\\

\hline

    \end{tabular}
    \label{appendix:graph}
\end{table*}

\end{document}